# MaskLRF: Self-supervised Pretraining via Masked Autoencoding of Local Reference Frames for Rotation-invariant 3D Point Set Analysis

Takahiko Furuya[1]

[1]Integrated Graduate School of Medicine, Engineering, and Agricultural Sciences, University of Yamanashi, 4-3-11 Takeda, Kofu-shi, Yamanashi-ken, 400-8511, Japan

Corresponding author: Takahiko Furuya (e-mail: takahikof@yamanashi.ac.jp).

This work was supported by the Japan Society for the Promotion of Science (JSPS) KAKENHI (Grant No. 21K17763, 24K14992).

**ABSTRACT** Following the successes in the fields of vision and language, self-supervised pretraining via masked autoencoding of 3D point set data, or Masked Point Modeling (MPM), has achieved state-of-the-art accuracy in various downstream tasks. However, current MPM methods lack a property essential for 3D point set analysis, namely, invariance against rotation of 3D objects/scenes. Existing MPM methods are thus not necessarily suitable for real-world applications where 3D point sets may have inconsistent orientations. This paper develops, for the first time, a rotation-invariant self-supervised pretraining framework for practical 3D point set analysis. The proposed algorithm, called MaskLRF, learns rotation-invariant and highly generalizable latent features via masked autoencoding of 3D points within Local Reference Frames (LRFs), which are not affected by rotation of 3D point sets. MaskLRF enhances the quality of latent features by integrating feature refinement using relative pose encoding and feature reconstruction using low-level but rich 3D geometry. The efficacy of MaskLRF is validated via extensive experiments on diverse downstream tasks including classification, segmentation, registration, and domain adaptation. The experiments demonstrate that MaskLRF achieves new state-of-the-art accuracies in analyzing 3D point sets having inconsistent orientations. Code will be available at: https://github.com/takahikof/MaskLRF .

## I. INTRODUCTION

Deep learning is an essential technique for accurate 3D point set analysis. Notably, in recent years, self-supervised pretraining of Deep Neural Networks (DNNs) for 3D point sets has become one of the hottest research topics in 3D vision [1], [2]. Self-supervised pretraining leverages a large amount of unlabeled 3D point sets instead of labeled ones, which are often difficult to collect due to high labeling costs. The typical framework of self-supervised pretraining first trains an encoder DNN, or backbone, via a pretext task using unlabeled 3D point sets as training data. The pretrained backbone is then finetuned for specific downstream tasks by using (usually small amount of) labeled 3D point sets. Accuracies for downstream tasks highly depend on the pretext task used for pretraining.

Following the successes of masked language modeling [3] and masked image modeling [4], [5], there has been a growing interest in masked autoencoding of 3D point sets, also referred to as Masked Point Modeling (MPM) [6]. MPM is a pretext task where a DNN reconstructs a set of erased, or masked, local regions from an incomplete input 3D point set consisting of unmasked 3D points. The recently proposed MPM methods ([6], [7], [8], [9], [10], [11]) employ Transformer [12] as a backbone DNN for its capability to refine local shape features considering their interrelationships. These MPM methods thus can acquire expressive latent 3D shape features that capture both local shape geometry and global shape context, leading to state-of-the-art accuracy in 3D point set analysis.

However, the existing MPM methods have a drawback; they are not invariant to SO(3) rotations of 3D point sets. The previous studies on MPM use 3D point sets whose orientations are consistently aligned by humans, at all the stages of pretraining, finetuning, and evaluation. I argue that such a strongly constrained setup is not always practical since orientations of 3D point sets are generally inconsistent in real-world application scenarios. For example, the orientations of scanned real-world 3D objects can vary depending on the poses of both an object and a range scanner. Or, the correspondence between the upright direction of a synthetic 3D shape and one of the three axes of a coordinate system depends on 3D modeling software. The existing MPM methods thus end up in limited use cases.

This paper aims at developing a practical and versatile self-supervised pretraining framework for 3D point set analysis. To this end, I propose a novel rotation-invariant (RI) MPM algorithm called *MaskLRF* (Fig. 1). The core idea of MaskLRF is simple; It normalizes the orientation of each local region of a 3D point set by using Local Reference Frame (LRF) [4], [13], and performs masked autoencoding on a set of rotation-normalized local regions.

However, designing a Transformer-based RI MPM framework is non-trivial due to the following two issues. First, I cannot use (absolute) positional encoding [6], which is not only essential for feature refinement but also serves as a "prompt" for 3D shape reconstruction. The positional encoding used by the existing MPM methods is a rotation-covariant quantity that changes with rotation of 3D points, and thus is not suitable for RI MPM. Second, a proper reconstruction target is not clear since there is no prior work on RI MPM. Some recent studies on non-RI MPM [11], [14], [15] found that reconstructing weakly encoded low-level features instead of raw 3D points leads to better latent features. In light of these findings, the reconstruction target for RI MPM should also be chosen carefully.

To address the first issue, I propose *relative pose encoding*, which describes both relative position and relative orientation among local regions of a 3D point set. The relative pose encoding is an RI quantity. Therefore, it realizes MPM of 3D point sets having inconsistent orientations. For the second issue, I assume that the finding by the non-RI MPM algorithms [11], [14], [15] is also valid for RI MPM. That is, reconstructing features that describe low-level but rich 3D geometry of rotation-normalized local regions will enhance the quality of latent features. Based on this assumption, MaskLRF employs a handcrafted shape feature having 3D grid structure as the reconstruction target.

The effectiveness of MaskLRF is verified on various downstream tasks including real-world object classification, few-shot object classification, part segmentation, scene registration, and domain adaptation. The experiments show that MaskLRF achieves new state-of-the-art accuracies in analyzing 3D point sets having inconsistent orientations.

Contributions of this paper are summarized as follows.

- Developing a first-of-its-kind self-supervised pretraining framework specialized for *rotation-invariant* (RI) analysis of 3D point sets.

- Proposing a novel Masked Point Modeling (MPM) framework called MaskLRF. It accepts rotationally inconsistent 3D point sets at all stages of pretraining, finetuning, and evaluation. MaskLRF thus extends potential use cases of the current Transformers for 3D point set, which do not have rotation invariance.

- Comprehensively evaluating the effectiveness of MaskLRF. It achieves accuracy higher than existing MPM methods and existing RI DNNs in analyzing 3D point sets having inconsistent orientations.

## II. RELATED WORK

### A. SELF-SUPERVISED PRETRAINING FOR 3D POINT SET ANALYSIS

Self-supervised pretraining enables DNNs to learn general-purpose feature representations which can be transferred to various downstream tasks. A number of self-supervised pretraining algorithms for 3D point set analysis have been proposed [1], [2]. Among them, promising methods can be categorized into two approaches, i.e., contrastive learning-based ([16], [17], [18], [19]) and MPM-based ([6], [7], [8], [9], [10], [11], [14], [15], [20], [21], [22]) approaches.

PointContrast by Xie et al. [16] is the pioneering work of the contrastive learning-based approach. [16] employs a pretext task that compares latent 3D shape features extracted from randomly augmented two 3D point set scenes. Zhang et al. [17] propose to use multi-view depth maps as training data for self-supervised pretraining. Depth maps are converted to two different shape representations (i.e., point sets and voxels) and are processed by a DNN for contrasting their latent shape features. Rao et al. [18] propose a contrastive learning using synthetic 3D point set scenes instead of using scanned 3D scenes. Long et al. [19] contrast the latent features extracted by a DNN with the prototype latent features acquired by clustering, both at the point-level and object-level.

The pretext task by Han et al. [22], called half-to-half reconstruction, can be viewed as an early generation MPM. In [22], a 3D object is split into two partial 3D point sets having nearly equal size and a DNN is trained by reconstructing one of the partial 3D point sets from the other partial 3D point set given as an input. Wang et al. [21] create incomplete 3D point sets by masking all the occluded 3D points when a synthetic 3D point set is viewed from a certain perspective. The DNN of [21] is trained so that it complements the occluded points of an incomplete 3D point set. These early generation MPM algorithms [21], [22], however, use an inadequately expressive backbone DNN such as PointNet [24] or its variant [23], which makes pretraining less effective.

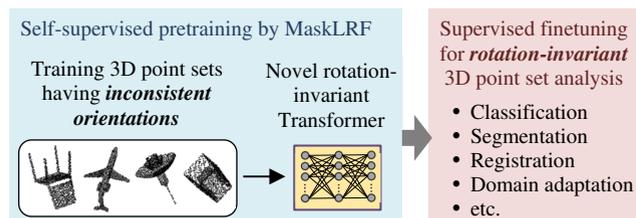

**FIGURE 1.** This paper proposes a self-supervised pretraining framework tailored to rotation-invariant (RI) analysis of 3D point sets. The parameters of the pretrained RI Transformer are used as initialization of finetuning for diverse downstream tasks.

To improve the MPM framework, Transformer model [12] was introduced as a new backbone DNN. The typical learning procedure for the Transformer-based MPM is as follows. Firstly, an input training 3D point set is split into multiple local regions and a certain percentage of the local regions are masked. The unmasked, or visible, regions are then described as feature vectors called tokens. Transformer takes as its input the set of visible tokens and refines the tokens by using the self-attention mechanism [12] that is capable of considering interrelationships among the tokens. Transformer is trained so that it reconstructs 3D point sets within masked local regions.

Point-BERT by Yu et al. [6] and Point-MAE by Pang et al. [8] are the first Transformer-based MPM algorithms that employ the abovementioned learning procedure. Point-BERT and Point-MAE have outperformed most of the existing contrastive learning-based methods and the early generation MPM methods in various downstream tasks. Therefore, the Transformer-based MPM can currently be positioned as the state-of-the-art approach for self-supervised pretraining for 3D point set analysis. More recent studies have attempted to improve the Transformer-based MPM by introducing, for example, hierarchical Transformer architecture [9], discriminative pretext task [7], and autoregressive generative pretext task [11]. As mentioned in Section I, however, all the existing Transformer-based MPM do not have invariance against SO(3) rotations of 3D point sets. They are thus not suitable for downstream tasks that require rotation invariance. In contrast, this paper is unique since it realizes RI Transformer-based MPM to extend potential use cases.

Apart from the abovementioned single-modal approaches that use only 3D point set data, some studies [25], [26], [27], [28], [29] have attempted to leverage knowledge of different modalities, e.g., 2D image and/or text. [28] and [29] employ a vision-language model (e.g., CLIP [30]) to improve both masking strategy and pretext task of the Transformer-based MPM. Note that the use of knowledge from different data domain is beyond the scope of this paper. My study falls into a single-modal Transformer-based MPM approach; I use only 3D point set data to obtain rotation-invariant and highly generalizable latent 3D shape features.

*B. ROTATION-INVARIANT 3D POINT SET ANALYSIS*

Rotation invariance is essential for practical 3D point set analysis. Various RI DNNs for 3D point sets have been proposed. They can be classified into the following three approaches; extracting inherently RI feature ([31], [32], [33], [34], [35], [36], [37]), designing rotation-equivariant DNN architecture ([38], [39]), and normalizing rotation of 3D point sets ([40], [41], [42], [43], [44], [45], [46], [47]).

**Extracting inherently RI feature.** The group of prior studies [31], [32], [33], [34], [35], [36], [37] have used 3D geometric features that are not affected by the rotation of input 3D point sets. These methods sample local regions of a 3D point set at the initial layer of a DNN. The local regions are encoded using RI low-level features, such as distances between 3D points and angles among surface normals. These RI local features are then propagated through subsequent layers to generate an object-level RI feature. While being inherently RI, encoding to low-level features such as distances and angles results in a significant loss of 3D shape information.

**Designing rotation-equivariant DNN architecture.** Shen et al. [38] and Deng et al. [39] have extend DNN neurons from 1D to 3D so that they can preserve SO(3) rotation of the input 3D point set. Such rotation-equivariant shape features are converted to RI features by computing inner product of two identical rotation-equivariant shape features [39] or taking the norms of 3D neurons [38]. However, as noted by [35], the rotation-equivariant DNNs need to impose strong constraints, such as linearity, on their layers to achieve rotation equivariance, sacrificing flexibility in feature extraction.

**Normalizing rotation.** The studies in this category achieve rotation invariance by normalizing the orientation of a 3D point set at global scale [37], [42] or local scale [40], [43], [47]. Compared to global scale, rotation normalization at local scale is easier since 3D shape in a local region tends to be simple. Furuya et al. [40] compute a local coordinate system called Local Reference Frame (LRF) to rotation-normalize each local region. The LRF of [40] is computed by applying Principal Component Analysis (PCA) to the 3D points within a local region. Specifically, three mutually orthogonal axes for rotation normalization are computed by eigendecomposition of the covariance matrix of the 3D points in a local region. Luo et al. [43] and Zhang et al. [47] propose a DNN block that predicts an intrinsic LRF of a local region for its rotation-normalization. The use of LRF achieves rotational invariance without loss of 3D shape information in a local region. I thus considered LRF to be a suitable means for RI MPM.

All the existing RI DNNs described above assume to be supervisedly trained from scratch without pretraining. Self-supervised pretraining of RI DNNs has not yet been studied. Spezialetti et al. [48] and Kim et al. [49] proposed self-supervised pretext tasks that predict canonical orientation of 3D point sets. However, these methods are not rigorously rotation invariant since they employ non-RI DNNs. Recently, Spezialetti et al. [50] and Furuya et al. [51] introduced self-supervised learning of fully rotation-invariant 3D point set features. However, their methods are designed not for pretraining, but for a specific task such as registration or retrieval. In contrast, my method is versatile; it acquires highly generalizable RI features useful for various downstream tasks.

## III. PROPOSED ALGORITHM

### A. OVERVIEW OF PROPOSED ALGORITHM

Section III elaborates MaskLRF, which bridges the state-of-the-art self-supervised pretraining (i.e., Transformer-based MPM) to rotation-invariant 3D point set analysis. Fig. 2 illustrates the overview of MaskLRF. Hereafter, a local region cropped from a 3D point set is called a patch. A training 3D point set is first divided into visible patches and masked patches, each of which is associated with an LRF. The visible patches are rotation-normalized and then embedded as token features by the DNN layers. For token refinement, I design *Relative pose-aware Rotation-invariant Point Transformer (R2PT)*. R2PT has three differences compared to the Transformers used in the previous MPM methods. (i) R2PT incorporates relative pose encoding into the attention computation to consider mutual pose differences between rotation-normalized patches. (ii) R2PT effectively and efficiently captures both local geometry and global context of a 3D point set by alternately applying local attention and global yet sparse attention. (iii) The entire DNN is effectively pretrained via the pretext task that involves reconstruction of handcrafted 3D grid-structured features extracted from masked patches. After pretraining by MaskLRF, the encoder part of R2PT is finetuned for downstream tasks.

**Motivation for using LRF:** Among the various approaches to rotation-invariance reviewed in Section 2, this study adopts LRF, which has been well-studied in the literature [52], [53], [54]. The computation of LRF is relatively lightweight as it is obtained by eigenvalue decomposition of the covariance matrix of 3D points within a patch. In addition, unlike inherently RI low-level features, LRF preserves the distribution of 3D points within a patch as-is. I expect that such a lossless input representation realizes accurate feature refinement by the Transformer.

### B. PATCHIFICATION AND MASKING

MaskLRF employs a patchification procedure similar to the existing MPM methods [6], [8]. A training 3D point set $\mathbf{X}$ consists of $n$ oriented 3D points where each 3D point is associated with its normal vector. The center 3D points of $N_p$ patches are obtained by applying Farthest Point Sampling (FPS) [23] to $\mathbf{X}$. Each patch is formed by collecting $k$ 3D points closest to its center point $\mathbf{c}$. A set of 3D points within a patch is represented as a matrix $\mathbf{S} \in \mathbb{R}^{k \times 3}$. To obtain invariance against translation, 3D points within $\mathbf{S}$ are

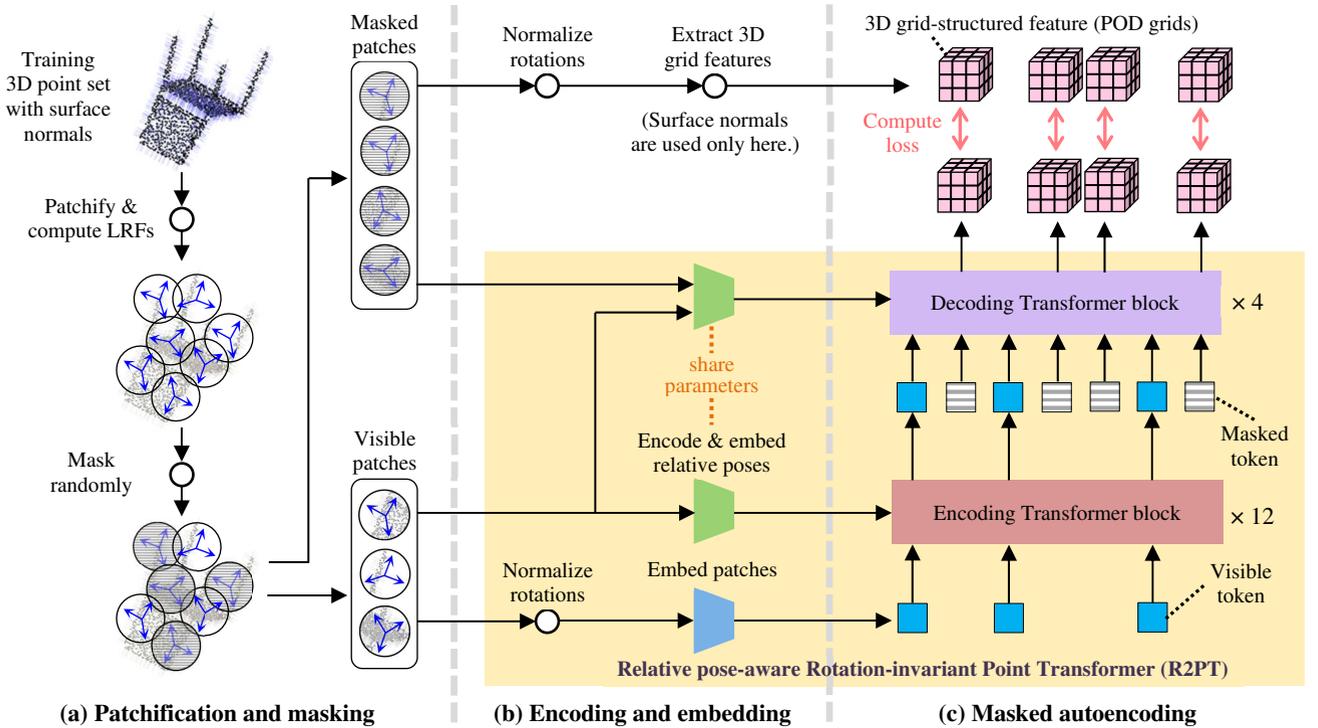

**(a) Patchification and masking** | **(b) Encoding and embedding** | **(c) Masked autoencoding**

**FIGURE 2.** The overview of MaskLRF. (a) A training 3D point set is divided into multiple patches, each of which is associated with its Local Reference Frame (LRF). A certain percentage (e.g., 60%) of the patches is randomly masked. (b) Every visible/masked patch is rotation-normalized by using its LRF. Each visible patch is then embedded by the DNN layers while each masked patch is described by a shape feature having 3D grid structure. Relative poses among the patches are encoded by using their positions and LRFs. (c) The entire DNN is trained via the masked autoencoding task where the DNN is forced to reconstruct the 3D grid-structured features of the masked patches. The Transformer blocks effectively refine the tokens by simultaneously considering their feature similarities and relative poses.

represented by normalized coordinates with respect to its center point **c**. Throughout this paper, $n$, $N_p$, and $k$ for pretraining are fixed at 1,024, 64, and 32, respectively. The patches may thus spatially overlap each other.

For each patch, MaskLRF computes an LRF, which consists of three mutually orthogonal axes $\mathbf{e}_1$, $\mathbf{e}_2$, and $\mathbf{e}_3$. In this paper, the first axis $\mathbf{e}_1$ is obtained by applying PCA to the covariance matrix of **S**. $\mathbf{e}_1$ corresponds to the eigenvector associated with the smallest eigenvalue. The sign of $\mathbf{e}_1$ is disambiguated by using the method in [55]. $\mathbf{e}_1$ estimates the normal of an object surface. Note that the surface normal associated with **X** is not used as $\mathbf{e}_1$, so that MaskLRF can process 3D point sets without normals in finetuning and evaluation. The second axis is obtained as in [35]. That is, $\mathbf{e}_2$ is computed by projecting the vector from the center point **c** to the barycenter of the patch onto the plane perpendicular to $\mathbf{e}_1$. The third axis $\mathbf{e}_3$ is computed as the cross product of $\mathbf{e}_1$ and $\mathbf{e}_2$. The LRF is represented as a 3×3 matrix **F**, whose column corresponds to one of $\mathbf{e}_1$, $\mathbf{e}_2$, and $\mathbf{e}_3$. As a result, each patch $P$ is represented as a triplet (**S**, **c**, **F**).

For masking, MaskLRF employs the same strategy as in [6], [8]. Specifically, $M$ % of $N_p$ patches are randomly chosen as masked patches, and the remaining patches are treated as visible patches. In the experiments, the masking ratio $M$ during pretraining is set to 60. The influence of $M$ on the accuracy of a downstream task is evaluated in the experiments. At the finetuning stage, masking is not performed. That is, all the $N_p$ patches are treated as visible ones.

### C. RELATIVE POSE-AWARE ROTAION-INVARIANT POINT TRANSFORMER (R2PT)

**Relative pose encoding and its embedding.** Relative pose encoding attempts to compensate for the loss of pose information caused by normalizing rotations of the patches. In the encoder block of R2PT, a relative pose encoding is computed for every pair of visible patches cropped from the 3D point set **X**. The decoder block, on the other hand, computes relative pose encodings among all the visible/masked patches of **X**. A relative pose encoding comprises two quantities, i.e., relative position and relative orientation. Let a pair of two patches be $P_i = (\mathbf{S}_i, \mathbf{c}_i, \mathbf{F}_i)$ and $P_j = (\mathbf{S}_j, \mathbf{c}_j, \mathbf{F}_j)$. The relative position $\text{RP}_{ij}$ and relative orientation $\text{RO}_{ij}$ of $P_i$ with respect to $P_j$ are calculated by Eq. 1 and 2, respectively.

$$\text{RP}_{ij} = (\mathbf{c}_i - \mathbf{c}_j)\mathbf{F}_j, \quad \text{RP}_{ij} \in \mathbb{R}^3 \quad (1)$$

$$\text{RO}_{ij} = \mathbf{F}_i^T \mathbf{F}_j, \quad \text{RO}_{ij} \in \mathbb{R}^{3\times 3} \quad (2)$$

$\text{RP}_{ij}$ denotes the position of $\mathbf{c}_i$ in the LRF $\mathbf{F}_j$ whose origin is $\mathbf{c}_j$, while $\text{RO}_{ij}$ corresponds to the rotation matrix that aligns $\mathbf{F}_i$ with $\mathbf{F}_j$ [55]. $\text{RP}_{ij}$ and $\text{RO}_{ij}$ are constant regardless of any rotation of **X**. $\text{RP}_{ij}$ and $\text{RO}_{ij}$ are then flattened and concatenated to obtain a 12D relative pose encoding vector.

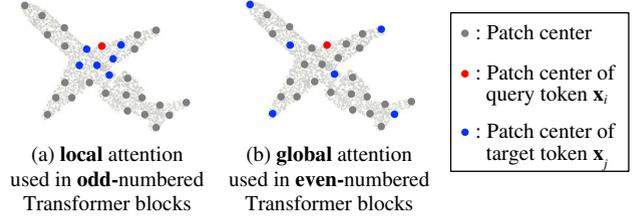

**FIGURE 3. R2PT effectively and efficiently captures both local shape geometry and global shape context of a 3D point set by alternately applying the local and global self-attention.**

The 12D vector is embedded in higher-dimensional space by using the two-layer MLP. The parameters of this MLP are shared across all the encoding/decoding Transformer blocks in R2PT. The relative pose embedding, denoted as $\mathbf{R}_{ij} \in \mathbb{R}^d$ ($d = 384$), is used for the self-attention computation described below.

**Encoder part.** Each visible patch $P_i$ is rotation-normalized by computing $\mathbf{S}_i \mathbf{F}_i$. The patch is then converted to a token feature $\mathbf{x}_i \in \mathbb{R}^d$ by the PointNet [24]-like DNN, as in [6], [7], and [8]. The encoder part is a series of 12 Transformer blocks. Each block takes as input a set of tokens $(\mathbf{x}_1, ..., \mathbf{x}_{N_v})$ and outputs a set of refined tokens $(\mathbf{y}_1, ..., \mathbf{y}_{N_v})$ where $\mathbf{y}_i \in \mathbb{R}^d$. $N_v$ is the number of visible tokens, which is equal to $(1-M/100)\times N_p$ for pretraining and $N_p$ for finetuning.

A typical self-attention requires a spatial and temporal complexity of $O(N_v^2)$. In addition to attention computation, MaskLRF requires the same complexity for relative pose embedding. The encoder works efficiently during pretraining since $N_v$ is small. In contrast, however, finetuning suffers from high complexity. To reduce the cost, I propose to subsample attention target tokens. This paper employs two types of self-attention, i.e., local attention to capture local shape geometry (Fig. 3a) and global attention to capture long-range shape context (Fig. 3b). For each query token $\mathbf{x}_i$, local attention collects, in the 3D space, $t$ nearest neighbors as attention targets for $\mathbf{x}_i$. Global attention obtains $t$ attention targets for $\mathbf{x}_i$ by applying FPS to the patch center points. Inspired by [57] for 2D image analysis, local attention is used in the odd-numbered Transformer blocks and global attention is used in the even-numbered Transformer blocks. Such an alternate block arrangement is expected to progressively refine the tokens considering both local geometry and global context at a low computation cost. For pretraining, $t$ is set to $N_v/4$. $t$ for finetuning is fixed at 16 regardless of $N_p$, which varies depending on a downstream task.

After subsampling the attention targets, each query token $\mathbf{x}_i$ is refined by the relative pose-aware self-attention:

$$\mathbf{y}_i = \sum_{j \in \varphi(\mathbf{x}_i)} \alpha_{ij} \mathbf{v}_{ij} \quad (3)$$

In Eq. 3, $\varphi(\mathbf{x}_i)$ denotes a set of indices for the attention targets of $\mathbf{x}_i$. $\alpha_{ij}$ and $\mathbf{v}_{ij}$ are the attention score and value vector between tokens $\mathbf{x}_i$ and $\mathbf{x}_j$, respectively. Note that not only $\alpha_{ij}$ but also $\mathbf{v}_{ij}$ reflects the relation between the tokens. $\alpha_{ij}$ and $\mathbf{v}_{ij}$ are computed by the following equations.

$$\mathbf{v}_{ij} = \mathbf{x}_j \mathbf{W}^V + \mathbf{R}_{ij} \qquad (4)$$

$$\alpha_{ij} = \exp(e_{ij}) \Big/ \sum_{l \in \varphi(\mathbf{x}_i)} \exp(e_{il}) \qquad (5)$$

$$e_{ij} = \frac{(\mathbf{x}_i \mathbf{W}^Q)(\mathbf{x}_j \mathbf{W}^K)^T + (\mathbf{x}_i \mathbf{W}^Q)\mathbf{R}_{ij}^T}{\sqrt{d}} \qquad (6)$$

In the equations, the terms colored in green hold relative pose information. Omitting the green terms boils down to the original self-attention [12]. $\mathbf{W}^Q$, $\mathbf{W}^K$, $\mathbf{W}^V \in \mathbb{R}^{d \times d}$ are linear projections with learnable parameters. I use six heads for multi-head self-attention. The set of refined tokens is further processed by an MLP with skip connection as in [12], and is fed into the subsequent Transformer block.

**Decoder part.** The decoder part comprises four Transformer blocks. Each block receives $N_v$ visible tokens and $N_m$ masked tokens ($N_m = N_p - N_v$). The initial masked token, which is a $d$-dimensional learnable vector, is duplicated $N_m$ times and input to the first decoding block. Each decoding block refines a total of $N_p$ tokens by using the relative-pose aware self-attention as in the encoder part. The refined masked tokens from the last decoding block are further processed by a single-layer prediction head to reconstruct the set of feature vectors denoted as ($\mathbf{z}_1$, ..., $\mathbf{z}_{Nm}$).

### D. OBJECTIVE OF PRETRAINING

**Reconstruction target.** Most MPM methods employ a pretext task that needs to reconstruct raw 3D point sets within masked patches. More recent studies found that reconstructing weakly encoded features, such as surface variations [20] or binary occupancy grids [14], [15], leads to effective pretraining. This paper proposes to reconstruct more expressive 3D geometric features. The reconstruction target of MaskLRF is represented as a 3D grid-structured feature, which describes low-level but rich 3D geometry. Specifically, the bounding box of each rotation-normalized masked patch is spatially partitioned by 6×6×6 regular grids. Each grid cell is described by a 10D feature called POD [58], which consist of the frequency (1D), the mean coordinates (3D), and the covariance of normal vectors (6D). The 3D grid feature is then flattened to obtain a feature vector $\hat{\mathbf{z}}_i$ having 6×6×6×10 = 2160D. Note that the surface normals of a training sample are only used to extract its POD features.

**Intuition behind POD grid reconstruction.** MaskLRF does not reconstruct raw 3D point sets since shapes of rotation-normalized patches are less diverse compared to those of non-normalized patches used by the non-RI MPM methods. In the non-RI MPM methods, even simple patches (e.g., rods or planes) can have various orientations, making diverse local shapes available for pretraining. In contrast, patches of MaskLRF are less diverse since the 3D points in a patch are rotation-normalized by using the fixed rule described in Section III.B. In this case, the DNN can easily reduce the loss for point set reconstruction by generating simple (e.g., elliptical or planar) shapes having a specific orientation. As a result, the pretraining may converge to a suboptimal solution. This paper thus proposes the POD grid reconstruction to effectively train the DNN using patches with limited shape diversity. The POD grid reconstruction imposes the following challenging pretext task on the DNN: "Are cells within a 3D grid occupied by points? If so, predict the 3D geometric features within that cell". In other words, the proposed reconstruction task requires jointly solving two predictive tasks, i.e., occupancy prediction and geometric feature prediction, which facilitate learning of better latent shape features.

**Loss and optimization.** The loss for each training 3D point set is calculated by Eq. 7.

$$L = \sum_{i=1}^{N_m} \| \mathbf{z}_i - \hat{\mathbf{z}}_i \|_2^2 \qquad (7)$$

AdamW [59] is used for optimization. The learning rate is initialized at $10^{-3}$ and is decreased to $10^{-6}$ by using a cosine scheduling. Pretraining is iterated for 300 epochs with the batch size of 64. During pretraining (and also in finetuning), each training sample is augmented by random anisotropic scaling with scaling ratios ranging from 0.8 to 1.2.

### E. FINETUNING

After pretraining, the decoder part of R2PT is replaced with a task-specific prediction head whose parameters are initialized randomly. The parameters of the entire DNN are then finetuned. In the process of prediction, each task-specific prediction head computes different types of 3D shape features, i.e., global features and pointwise features, depending on the downstream task.

Global features are mainly used for classification tasks. Each encoding block of R2PT is expected to capture shape features at different semantic levels due to the alternate local/global block arrangement. All refined tokens are thus used for a global feature. Specifically, at each of the 12 encoding blocks, output tokens are aggregated by average pooling. The resultant 12 aggregated tokens are then concatenated to create a global feature per 3D point set.

Pointwise features are used for tasks that require per-point prediction, such as segmentation and registration. The token features output from the last encoding Transformer block are upsampled to pointwise features. This paper adopts Pose-aware Feature Propagation devised in [47] for upsampling. This module leverages not only distances among neighboring 3D points but also angles among LRFs.

The finetuning also uses AdamW. The learning rate increases linearly from 0 to $\eta$ during the first 10 epochs. After the 10-th epoch, it decreases toward 0 by using a cosine scheduling. The hyperparameters used for finetuning are presented in Section IV-A.

## IV. EXPERIMENTS AND RESULTS

### A. EXPERIMENTAL SETUP

The effectiveness of MaskLRF is verified on five downstream tasks of 3D point set analysis, i.e., real-world object classification, few-shot object classification, part segmentation, scene registration, and domain adaptation.

**Competitors.** The experiments on the classification and segmentation tasks use existing self-supervised pretraining methods as competitors. They are six MPM-based methods [6], [7], [8], [9], [10], [11]. In addition, previously proposed DNNs having rotation invariance [35], [36][1], [37], [39], [40], [42], [43], [45], [47], [60] are included in the set of competitors. These RI DNNs are not pretrained, but are trained from scratch for each downstream task. Furthermore, I add pretrained RI DNNs to the competitors for a fair evaluation. Specifically, the two state-of-the-art RI DNNs [36], [47] are pretrained by using the state-of-the-art self-supervised representation learning algorithm for 3D point set called SDMM [51]. SDMM is built upon the self-distillation framework [61] originally proposed for 2D images. On the other hand, the experiments on the scene registration and domain adaptation tasks compare MaskLRF with existing methods specifically designed for each task.

**Dataset for pretraining.** For a fair comparison, all the pretraining methods, including MaskLRF, use the ShapeNetCore55 dataset [62] for self-supervised pretraining. ShapeNetCore55 contains over 50,000 3D shapes categorized into 55 semantic classes. Note that the category labels are not used for self-supervised pretraining. Each 3D point set consists of $n = 1,024$ points.

**Rotation settings.** To evaluate rotation invariance of the methods, the experiments use three rotation settings, i.e., A/A, A/R, and R/R. "A" stands for "consistently aligned", while "R" means "randomly rotated". The A/A setting uses 3D point sets whose orientations are consistently aligned by humans at all the stages of pretraining, finetuning, and evaluation. A/A is the easiest setting, but would not fit to real-world application scenarios. In the A/R setting, pretraining and finetuning use consistently aligned 3D point sets while evaluation uses randomly rotated 3D point sets. R/R uses randomly rotated 3D point sets throughout pretraining, finetuning, and evaluation. RI methods should produce similar accuracy values in the three rotation settings. Note that identical accuracy values are not expected even for RI methods due to randomness in DNN training.

**Hyperparameters for finetuning.** Table I shows the hyperparameter values used for each downstream task. For the number of epochs for classification and segmentation, I follow the evaluation protocol of the existing studies [6], [8]. The number of patches $N_p$ is set larger for the segmentation and registration tasks since they require pointwise dense prediction.

TABLE I. Hyperparameters of MaskLRF used for finetuning. ($b$: batch size, $e$: number of epochs, $\eta$: initial learning rate, $N_p$: number of patches per 3D shape, $k$: number of 3D points contained in a patch, $t$: number of attention targets per token).

| Downstream tasks | $b$ | $e$ | $\eta$ | $N_p$ | $k$ | $t$ |
|---|---|---|---|---|---|---|
| Real-world object classification | 32 | 300 | $5\times10^{-5}$ | 128 | 32 | 16 |
| Few-shot object classification | 32 | 150 | $5\times10^{-4}$ | 128 | 32 | 16 |
| Part segmentation | 24 | 300 | $5\times10^{-5}$ | 256 | 32 | 16 |
| Scene registration | 1 | 20 | $5\times10^{-5}$ | 1,024 | 32 | 16 |
| Domain adaptation | 32 | 30 | $5\times10^{-6}$ | 128 | 32 | 16 |

### B. COMPARISON AGAINST EXISTING ALGORITHMS

**Real-world object classification.** I use two benchmark datasets consisting of scanned 3D objects, i.e., ScanObjectNN (SONN) [63] and OmniObject3D (OO3D) [64]. SONN consists of 2,890 indoor objects classified into 15 categories. I use the official train/test split with 2,309 training shapes and 581 testing shapes. This paper reports accuracies for the three subsets of SONN, i.e., OBJ_BG, OBJ_ONLY, and PB_T50_RS. The OO3D dataset consists of 5,382 3D point sets classified into 216 diverse categories. Since no official train/test split is provided, I select roughly 80% of the dataset for training and the rest for testing, resulting in 4,219 training shapes and 1,163 testing shapes.

Table II demonstrates the effectiveness of MaskLRF. For all the four datasets, the existing MPM algorithms suffer from significant accuracy drop especially under the A/R setting. This is because the previous MPM framework does not have rotation invariance. The previously proposed RI DNNs exhibit rotation invariance, but their accuracies are inferior to MaskLRF. Pretraining of the previous RI DNNs performs better than training from scratch, but its improvement is marginal. In contrast, MaskLRF yields better accuracies, indicating that it acquires expressive RI features during pretraining with the help of relative pose encoding and 3D grid feature reconstruction. Interestingly, MaskLRF outperforms the existing MPM methods even under the A/A setting. This is probably because the orientations of 3D point sets in the datasets are not perfectly aligned. MaskLRF having rotation invariance is advantageous in that it can avoid the problem of orientation misalignment.

**Few-shot object classification.** I follow the evaluation protocol in [65]. [65] adopts "$K$-way $N$-shot" classification scenarios. It randomly selects $K$ classes from the ModelNet40 dataset [66], and then ($N$+20) samples are randomly chosen for each class. A DNN is trained on $K \times N$ samples and tested on the remaining $K \times 20$ samples. I conduct 10 independent experiments and report their average classification accuracy as in [65].

---
[1] Since the official code of RIConv++ [36] did not exhibit rotation invariance, I modified it to be rotation-invariant.

TABLE II. Accuracies [%] for the real-world 3D object classification task. (Pre.: self-supervised pretraining, RI: rotation invariance)

| Methods | Pre. | RI | SONN OBJ_BG dataset | | | SONN OBJ_ONLY dataset | | | SONN PB_T50_RS dataset | | | OO3D dataset | | |
|---|---|---|---|---|---|---|---|---|---|---|---|---|---|---|
| | | | A/A | A/R | R/R | A/A | A/R | R/R | A/A | A/R | R/R | A/A | A/R | R/R |
| PointBERT [6] | ✓ | | 87.4 | 32.4 | 86.1 | 88.1 | 29.2 | 86.0 | 83.1 | 23.9 | 81.7 | 70.9 | 9.3 | 70.4 |
| MaskPoint [7] | ✓ | | 89.3 | 29.0 | 87.6 | 88.1 | 29.1 | 85.9 | 84.3 | 25.4 | 82.9 | 72.8 | 9.1 | 70.5 |
| Point-MAE [8] | ✓ | | 90.2 | 26.5 | 88.3 | 88.2 | 29.9 | 87.7 | 85.2 | 24.7 | 84.4 | 71.4 | 9.1 | 70.5 |
| Point-M2AE [9] | ✓ | | 91.2 | 32.8 | 86.3 | 88.8 | 32.5 | 85.7 | 86.4 | 28.8 | 80.7 | 72.1 | 13.4 | 69.4 |
| MaskSurf [10] | ✓ | | 91.2 | 29.0 | 89.2 | 89.2 | 31.2 | 87.6 | 85.8 | 26.1 | 84.6 | 72.3 | 9.3 | 70.7 |
| PointGPT [11] | ✓ | | 91.6 | 31.5 | 85.4 | 90.0 | 31.6 | 85.2 | **86.9** | 25.5 | 80.6 | 71.7 | 9.7 | 57.9 |
| DLAN [40] | | ✓ | 82.6 | 82.9 | 82.8 | 82.2 | 82.5 | 82.0 | 74.9 | 75.0 | 74.9 | 66.3 | 66.5 | 66.3 |
| PoseSelector [42] | | ✓ | 80.6 | 80.5 | 80.9 | 82.3 | 81.4 | 81.9 | 76.1 | 75.8 | 75.8 | 63.7 | 63.8 | 64.0 |
| VN-DGCNN [39] | | ✓ | 69.8 | 70.1 | 69.7 | 71.1 | 70.6 | 72.1 | 64.3 | 64.7 | 65.3 | 51.9 | 52.1 | 51.6 |
| LGR-Net [37] | | ✓ | 85.1 | 85.2 | 85.3 | 85.2 | 85.7 | 85.6 | 77.1 | 76.6 | 76.6 | 69.0 | 68.7 | 69.0 |
| EOMP [43] | | ✓ | 75.6 | 75.3 | 75.8 | 75.6 | 76.4 | 75.8 | 67.4 | 69.2 | 68.0 | 53.6 | 54.3 | 54.3 |
| PaRI-Conv [35] | | ✓ | 87.4 | 87.3 | 86.6 | 84.3 | 84.8 | 84.7 | 82.3 | 82.3 | 82.4 | 68.1 | 68.2 | 67.9 |
| RIConv++ [36] | | ✓ | 89.7 | 90.0 | 89.7 | 88.8 | 88.3 | 88.0 | 85.4 | 85.2 | 85.3 | 71.2 | 70.9 | 71.1 |
| PaRot [47] | | ✓ | 88.2 | 88.5 | 88.0 | 85.3 | 85.3 | 85.8 | 82.3 | 82.5 | 82.6 | 70.9 | 70.8 | 70.9 |
| RIConv++ & SDMM | ✓ | ✓ | 90.2 | 90.4 | 90.2 | 88.8 | 89.0 | 88.6 | 85.3 | 85.2 | 84.8 | 71.5 | 71.7 | 71.3 |
| PaRot & SDMM | ✓ | ✓ | 90.7 | 90.7 | 90.6 | 88.4 | 88.8 | 88.7 | 83.6 | 83.8 | 83.6 | 73.1 | 73.3 | 72.5 |
| MaskLRF (proposed) | ✓ | ✓ | **93.1** | **93.1** | **93.3** | **90.2** | **90.1** | **89.8** | 86.7 | **86.7** | **86.8** | **76.5** | **76.6** | **76.8** |

TABLE III. Accuracies [%] for the few-shot 3D object classification task using the ModelNet40 few-shot dataset. (Pre.: self-supervised pretraining, RI: rotation invariance)

| Methods | Pre. | RI | 5-way 10-shot | | | 5-way 20-shot | | | 10-way 10-shot | | | 10-way 20-shot | | |
|---|---|---|---|---|---|---|---|---|---|---|---|---|---|---|
| | | | A/A | A/R | R/R | A/A | A/R | R/R | A/A | A/R | R/R | A/A | A/R | R/R |
| PointBERT [6] | ✓ | | 94.6 | 53.0 | 91.5 | 96.3 | 49.7 | 95.3 | 91.0 | 32.0 | 85.9 | 92.7 | 33.7 | 90.9 |
| MaskPoint [7] | ✓ | | 95.0 | 45.0 | 90.4 | 97.2 | 48.3 | 95.6 | 91.4 | 29.4 | 84.8 | 93.4 | 29.5 | 89.5 |
| Point-MAE [8] | ✓ | | 96.3 | 52.6 | 87.7 | 97.8 | 55.0 | 93.7 | 92.6 | 33.6 | 80.2 | 95.0 | 33.3 | 88.4 |
| Point-M2AE [9] | ✓ | | **96.8** | 58.0 | 89.1 | 98.3 | 58.1 | 94.5 | 92.3 | 39.2 | 83.6 | 95.0 | 38.6 | 89.5 |
| MaskSurf [10] | ✓ | | 96.5 | 50.9 | 90.9 | 98.0 | 49.5 | 95.0 | **93.0** | 33.0 | 83.3 | **95.3** | 31.6 | 90.2 |
| PointGPT [11] | ✓ | | **96.8** | 48.2 | 84.3 | **98.6** | 49.8 | 93.8 | 92.6 | 30.2 | 79.6 | 95.2 | 28.2 | 87.2 |
| DLAN [40] | | ✓ | 90.7 | 91.1 | 90.5 | 95.2 | 94.8 | 94.6 | 84.8 | 85.2 | 85.0 | 90.2 | 90.1 | 90.5 |
| PoseSelector [42] | | ✓ | 83.4 | 79.2 | 82.7 | 90.1 | 85.3 | 89.7 | 78.0 | 71.7 | 75.0 | 86.9 | 82.0 | 85.9 |
| VN-DGCNN [39] | | ✓ | 62.2 | 66.2 | 62.0 | 79.0 | 78.0 | 78.4 | 51.7 | 51.4 | 46.6 | 65.3 | 65.0 | 63.0 |
| LGR-Net [37] | | ✓ | 88.9 | 89.1 | 89.4 | 92.9 | 93.2 | 93.2 | 81.3 | 81.5 | 81.7 | 89.9 | 89.9 | 89.6 |
| EOMP [43] | | ✓ | 36.9 | 37.5 | 37.0 | 78.3 | 77.2 | 75.9 | 54.6 | 49.4 | 44.7 | 71.7 | 69.3 | 72.3 |
| PaRI-Conv [35] | | ✓ | 89.9 | 90.7 | 90.5 | 94.6 | 95.0 | 94.5 | 84.6 | 84.6 | 84.0 | 90.9 | 90.7 | 91.0 |
| RIConv++ [36] | | ✓ | 87.3 | 87.9 | 87.5 | 93.3 | 93.4 | 92.9 | 80.0 | 79.8 | 79.9 | 88.6 | 88.9 | 88.7 |
| PaRot [47] | | ✓ | 46.9 | 52.2 | 49.0 | 70.5 | 72.2 | 73.6 | 50.9 | 48.3 | 46.2 | 58.9 | 64.1 | 63.9 |
| RIConv++ & SDMM | ✓ | ✓ | 91.2 | 91.0 | 91.1 | 94.6 | 94.8 | 94.6 | 85.4 | 85.3 | 85.0 | 91.4 | 91.3 | 91.6 |
| PaRot & SDMM | ✓ | ✓ | 93.3 | 93.2 | 93.3 | 95.5 | 96.2 | 96.2 | 88.2 | 88.6 | 88.3 | 92.1 | 93.0 | 92.7 |
| MaskLRF (proposed) | ✓ | ✓ | 93.5 | **93.6** | **93.8** | 96.4 | **96.5** | **96.4** | 89.2 | **89.5** | **89.5** | 93.6 | **93.7** | **93.7** |

Table III shows accuracies for "5-way 10-shot", "5-way 20-shot", "10-way 10-shot", and "10-way 20-shot" scenarios. In Table III, MaskLRF yields superior classification accuracy over the existing RI DNNs. This result suggests that MaskLRF is better at mitigating overfitting during finetuning when labeled training samples are scarce. As in the real-world object classification experiment, the existing MPM methods show lower classification accuracy in the A/R and R/R settings compared to the A/A setting. This indicates that existing MPM methods have difficulty in acquiring rotation invariance in the few-shot classification scenario.

**Part segmentation.** I use the ShapeNetPart dataset [67], which contains 16,881 3D objects categorized into 16 semantic classes. I follow the evaluation protocol in [24]. That is, MaskLRF computes both global feature and pointwise feature described in Section III-E. Each pointwise feature is concatenated with the global feature and object class label, and then processed by an MLP to produce a pointwise prediction. Category-level mean Intersection-over-Union (C-mIoU) [24] is used as an accuracy index.

As shown in Table IV, the proposed MaskLRF clearly outperforms the existing methods under the A/R and R/R settings. This result implies that MaskLRF is capable of learning highly semantic pointwise features even when the orientations of the 3D shapes are inconsistent. Table V shows mean IoU for each object category. MaskLRF outperforms the existing methods in 12 out of 16 object categories.

TABLE IV. Accuracies (C-mIoU [%]) for the part segmentation task.

| Methods | Pre. | RI | ShapeNetPart dataset | | |
|---|---|---|---|---|---|
| | | | A/A | A/R | R/R |
| Point-MAE [8] | ✓ | | 84.2 | 38.7 | 79.4 |
| Point-M2AE [9] | ✓ | | **84.9** | 41.1 | 80.2 |
| MaskSurf [10] | ✓ | | 84.4 | 38.4 | 81.7 |
| PointGPT [11] | ✓ | | 84.1 | 38.8 | 79.4 |
| LGR-Net [37] | | ✓ | 80.0 | 80.0 | 80.1 |
| RMGnet [60] | | ✓ | – | 81.5 | 81.4 |
| AECNN [45] | | ✓ | 80.2 | 80.2 | 80.2 |
| RIConv++ [36] | | ✓ | – | 80.3 | 80.3 |
| PaRot [47] | | ✓ | – | 79.2 | 79.5 |
| RIConv++ & SDMM | ✓ | ✓ | 80.2 | 80.3 | 80.3 |
| PaRot & SDMM | ✓ | ✓ | 80.3 | 80.3 | 80.4 |
| MaskLRF (proposed) | ✓ | ✓ | 83.2 | **83.4** | **83.5** |

TABLE V. Category-wise mean IoU [%] for the part segmentation task under the R/R setting.

| Methods | aero | bag | cap | car | chair | earph. | guitar | knife | lamp | laptop | motor | mug | pistol | rocket | skate | table |
|---|---|---|---|---|---|---|---|---|---|---|---|---|---|---|---|---|
| Point-MAE [8] | 82.0 | 79.8 | 83.8 | 74.0 | 89.2 | 70.3 | 90.2 | **86.3** | 82.1 | 81.6 | 66.6 | 92.2 | 79.7 | 59.2 | 72.7 | 80.6 |
| Point-M2AE [9] | 82.9 | 79.9 | 82.4 | 75.0 | 89.6 | 72.8 | 90.9 | 85.0 | 83.4 | 83.7 | 71.0 | 93.7 | 80.8 | 57.6 | 74.1 | 81.1 |
| MaskSurf [10] | 83.0 | 79.5 | 84.6 | 76.8 | 89.9 | 74.7 | 91.1 | 85.6 | 83.7 | **89.1** | 72.0 | 92.5 | 82.6 | **66.1** | 74.5 | 80.6 |
| PointGPT [11] | 80.3 | 78.7 | 83.0 | 69.9 | 89.1 | 79.9 | 90.1 | 85.0 | 83.6 | 84.6 | 62.7 | 92.1 | 80.1 | 56.4 | 74.6 | 80.9 |
| LGR-Net [37] | 81.7 | 78.1 | 82.5 | 75.1 | 87.6 | 74.5 | 89.4 | 86.1 | 83.0 | 86.4 | 65.3 | 92.6 | 75.2 | 64.1 | **79.8** | 80.5 |
| RMGnet [60] | 82.4 | 81.0 | 85.7 | 76.9 | 89.7 | 79.7 | 91.5 | 84.1 | 81.9 | 84.7 | 72.6 | 93.8 | 81.9 | 61.4 | 77.5 | 79.5 |
| PaRot [47] | 82.9 | 82.1 | 83.2 | 75.7 | 89.4 | 76.1 | 91.5 | 86.1 | 81.4 | 80.3 | 59.3 | 94.3 | 79.7 | 57.0 | 73.3 | 79.2 |
| RIConv++ & SDMM | 83.1 | 81.0 | 87.7 | 77.5 | 90.3 | 65.5 | 90.8 | 84.8 | 82.9 | 82.3 | 69.0 | 94.3 | 79.1 | 54.7 | 73.7 | **81.3** |
| PaRot & SDMM | 83.9 | 77.8 | 86.0 | 75.9 | 89.3 | 74.0 | 91.3 | 85.5 | 82.4 | 81.2 | 68.7 | 93.9 | 78.6 | 57.7 | 75.1 | 80.1 |
| MaskLRF (proposed) | **85.2** | **84.5** | **88.9** | **80.0** | **90.5** | **80.6** | **91.6** | 84.8 | **84.6** | 86.9 | **76.7** | **95.4** | **82.8** | 64.6 | 77.7 | **81.3** |

**Scene registration.** Rotation invariance is essential for scene registration [68], [69]. As in [68], I use the rotated variant of the 3DMatch [70] and 3DLoMatch [71] datasets. These datasets include 62 indoor scenes, among which 46 are used for training, 8 for validation, and 8 for evaluation. A pair of scenes in 3DMatch and 3DLoMatch has >30% and 10-30% spatial overlap, respectively. As evaluation indices, I use Inlier Ratio (IR) [72] which measures an accuracy of point correspondence, and Registration Recall (RR) [70] that quantifies a success rate of registration. To adapt MaskLRF to the scene registration task, I replace the encoder and decoder of RoITr [69] with the pretrained encoder part of R2PT and the randomly initialized upsampling module mentioned in Section III-E, respectively. The entire DNN is then finetuned as in [69].

Table VI compares registration accuracies for rotated 3D point set scenes. As shown in Table VI, MaskLRF outperforms the existing DNN architectures specifically designed for scene registration. In particular, significant improvement in IR is observed. This result supports the claim in the part segmentation task; MaskLRF succeeds in obtaining highly semantic pointwise features.

TABLE VI. Accuracies (IR [%] and RR [%]) for registration of rotated 3D point set scenes.

| Methods | 3DMatch dataset | | 3DLoMatch dataset | |
|---|---|---|---|---|
| | IR | RR | IR | RR |
| YOHO [73] | 53.5 | 92.4 | 19.2 | 64.5 |
| RIGA [68] | 70.7 | 92.6 | 34.3 | 67.0 |
| GeoTrans [74] | 73.3 | 91.8 | 42.7 | 72.0 |
| RoITr [69] | 82.6 | 94.4 | 55.1 | 76.6 |
| MaskLRF (proposed) | **86.4** | **94.6** | **60.3** | **77.6** |

**Domain adaptation.** This subsection evaluates the capability of domain adaptation by MaskLRF. I use the adaptation scenario [75] from the synthetic 3D shape domain (source) to the real-world 3D shape domain (target). The source dataset consists of synthetic 3D point sets created from polygonal 3D shapes in the ModelNet10 (MN10) [66] or ShapeNet10 (SN10) [75] datasets. The target dataset consists of realistic 3D point sets in the ScanNet10 dataset [75] obtained by scanning real-world objects. After pretraining by MaskLRF, the proposed R2PT is trained to classify the object categories in the source dataset. Quality of domain adaptation is measured by classification accuracy for the target dataset. In addition to the synthetic/realistic domain gap, I observed that 3D shapes in the source and target domains had different upright orientations. Existing domain adaptation methods [75], [76], [77], [78], [79], [80] [81] preprocess the 3D shapes so that their upright axes are consistently aligned. However, such prior knowledge cannot be used for unknown datasets. Hence, I argue that rotation invariance is also essential for domain adaptation of 3D point sets.

Table VII compares classification accuracy on the target domain dataset. It is worth noting that MaskLRF which requires no upright orientation alignment produces the accuracy competitive to the existing methods that involve upright alignment. The result suggests that pretraining by MaskLRF is beneficial to domain adaptation, indicating high transferability of the learned RI features.

TABLE VII. Accuracies [%] for the domain adaptation task.

| Methods | MN10 → ScanNet10 | SN10 → ScanNet10 |
|---|---|---|
| PointDAN [75] | 44.8 | 45.7 |
| DefRec [76] | 42.6 | 46.1 |
| GAST [77] | 54.9 | 53.6 |
| MLSP [78] | 55.4 | 55.6 |
| ImplicitPCDA [79] | 55.3 | 55.4 |
| GLRV [80] | 53.6 | 49.1 |
| PC-Adapter [81] | **58.2** | 53.7 |
| MaskLRF (proposed) | 55.5 | **55.7** |

### C. IN-DEPTH EVALUATION OF MASK-LRF

This section verifies the effectiveness of each component of MaskLRF. The evaluation in this section uses classification accuracy [%] on the OBJ_BG subset of SONN and the OO3D dataset under the R/R rotation setting.

**Reconstruction target.** Table VIII compares reconstruction targets for pretraining. Evidently, the proposed POD grids lead to higher accuracy compared to the reconstruction targets devised in the previous studies. As mentioned in III-D, the proposed POD grid reconstruction imposes two challenging prediction tasks, i.e., occupancy prediction and geometric feature prediction. Jointly solving these two pretext tasks has a positive impact on the self-supervised pretraining of the proposed DNN. This is supported by the fact that the reconstruction of POD girds outperforms the reconstruction of occupancy grids [14], [15], which involves the occupancy prediction only. Table VIII also includes the results of the existing MPM methods whose reconstruction target is replaced with POD grids. The proposed POD grid reconstruction brings a slight but consistent improvement to pretraining by the existing MPM methods.

Fig. 4 exemplifies results of masked autoencoding by MaskLRF after pretraining. For visualization, raw 3D point sets are used as the reconstruction targets. Fig. 4 shows that MaskLRF successfully reconstructs the masked 3D points regardless of rotations of the input 3D shape, implying that MaskLRF acquires expressive RI latent features.

**Relative pose encoding.** I conduct an ablation study on the proposed relative pose encoding. In Table IX, "None" is the case that does not use relative pose information. That is, the green terms in Eq. 4 and 6 are omitted during pretraining and finetuning. "RP only" and "RO only" use either one of relative pose (Eq. 1) and relative orientation (Eq. 2). Table IX demonstrates that using both RP and RO contributes to effective pretraining. Without relative pose encoding, positional and orientational relationship among tokens cannot be used for feature refinement, resulting in significant accuracy drop.

TABLE VIII. Comparison of reconstruction targets for MPM.

| Methods | Reconstruction targets | OBJ_BG | OO3D |
|---|---|---|---|
| MaskLRF (proposed) | 3D points [6], [8], [9], [11] | 88.5 | 72.4 |
| | 3D points with normal [10] | 90.5 | 73.9 |
| | FPFH feature [11] | 90.4 | 73.4 |
| | Surface variations [20] | 72.6 | 49.9 |
| | Occupancy grids [14], [15] | 91.9 | 75.8 |
| | POD grids (proposed) | **93.3** | **76.8** |
| Point-MAE (A/A setting) | 3D points | 90.2 | 71.4 |
| | POD grids (proposed) | 90.5 | 73.3 |
| Point-M2AE (A/A setting) | 3D points | 91.2 | 72.1 |
| | POD grids (proposed) | 92.0 | 73.1 |

Table IX. Effectiveness of relative pose encoding.

| Relative pose encoding | OBJ_BG | OO3D |
|---|---|---|
| None | 81.3 | 65.2 |
| RP only | 93.1 | 76.0 |
| RO only | 87.6 | 73.0 |
| Both RP and RO (proposed) | **93.3** | **76.8** |

**Local/global self-attention.** Table X demonstrates the efficacy of combining local and global self-attention. Local attention alone fails to capture global context of 3D shape, while global attention alone has difficulty in describing local geometry. I also experiment a variant where the first half of the encoder/decoder part uses local attention only and the latter half uses global attention only. Table 8 shows that alternating local/global attentions enables effective feature refinement, resulting in high transferability to the downstream task.

Table X. Effectiveness of combining local/global attentions.

| Self-attention | OBJ_BG | OO3D |
|---|---|---|
| Local attention only | 90.1 | 74.0 |
| Global attention only | 90.3 | 74.2 |
| First half: local & latter half: global | 91.5 | 76.3 |
| Alternating local/global attention (proposed) | **93.3** | **76.8** |

**Amount of data for pretraining.** Table XI shows the influence of data amount for pretraining on the accuracy of the downstream task. In Table XI, "100%" uses all the ~50,000 samples of ShapeNetCore55 for pretraining. "0%" does not perform pretraining and the DNN is trained from scratch for the downstream task. Interestingly, even a pretraining with only 1% (~500) samples has a positive impact on the accuracy in the downstream task. This result indicates that the DNN architecture and pretext task of MaskLRF are appropriately designed for self-supervised pretraining. Table XI also shows that classification accuracy improves as the amount of data increases. The result implies that more (>50,000) data leads to more accurate latent 3D shape features. I leave pretraining using larger datasets and/or larger DNN models for my future work.

Table XI. Influence of amount of data for MaskLRF pretraining.

| Amount of data for pretraining | OBJ_BG | OO3D |
|---|---|---|
| 0 % (no pretraining) | 75.7 | 56.4 |
| 1 % | 83.1 | 70.5 |
| 5 % | 83.6 | 70.8 |
| 10 % | 89.4 | 72.9 |
| 50 % | 92.0 | 74.9 |
| 100 % (all samples in ShapeNetCore55) | **93.3** | **76.8** |

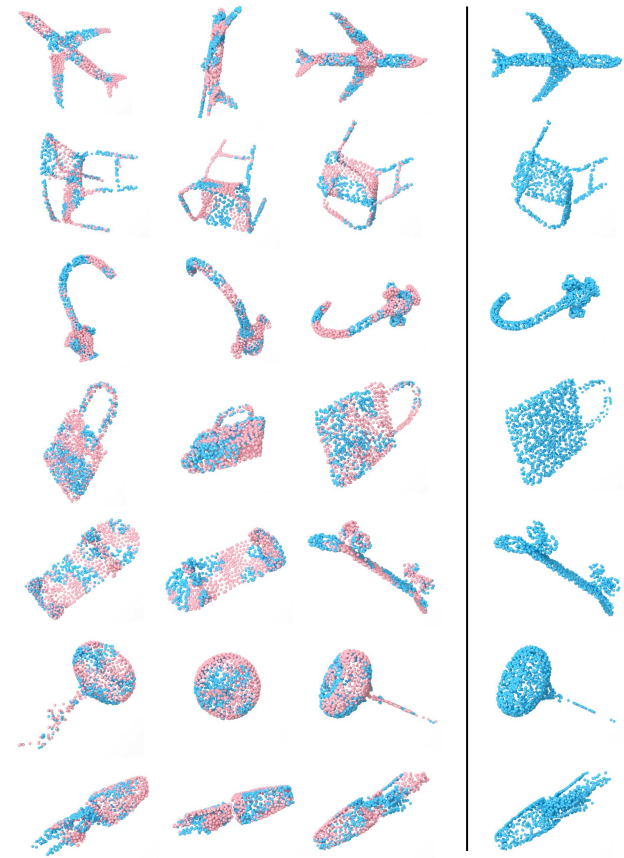

**FIGURE 4. Examples of masked autoencoding by MaskLRF.** In each row, the left three 3D point sets are reconstruction results for different orientations of the same 3D object. Visible (unmasked) points and predicted points are colored in blue and pink, respectively. The rightmost 3D point set is the groundtruth for its adjacent reconstruction result.

**Masking ratio.** Table XII shows the impact of the masking ratio $M$ during pretraining on the downstream task. The peak of the classification accuracy appears at around the masking ratio of 50-60%. These masking ratios are similar to those employed by the existing non-RI MPM methods (e.g., [8]). Decreasing $M$ approaches a simple autoencoding framework without masking. Using a too small $M$ (e.g., 10%) only forces the DNN to learn an identity mapping, which does not lead to learning of meaningful latent 3D shape features. On the other hand, using too large $M$ (e.g., 90%) makes the pretext task too difficult. Too few visible patches hamper the inference of an entire 3D shape to be reconstructed by the DNN, resulting in the low accuracy as shown in Table XII.

Table XII. Influence of the masking ratio $M$ for pretraining.

| Masking ratio $M$ | OBJ_BG | OO3D |
|---|---|---|
| 10 % | 92.1 | 74.8 |
| 20 % | 92.3 | 75.3 |
| 30 % | 92.8 | 76.1 |
| 40 % | 93.0 | 76.1 |
| 50 % | **93.4** | 76.6 |
| 60 % | 93.3 | **76.8** |
| 70 % | 92.8 | 75.4 |
| 80 % | 91.5 | 74.1 |
| 90 % | 88.8 | 73.4 |

**Computational efficiency.** Table XIII compares computational efficiency of the self-supervised pretraining methods. The experiment is done by using a PC having an *AMD Ryzen 9 5900X* CPU, 64GBytes of RAM, and an *NVIDIA GeForce RTX 3090* GPU with 24 GBytes of VRAM. Efficiency for finetuning is evaluated by the classification task on the SONN OBJ_BG dataset. The experiment includes two variants of MaskLRF, i.e., MaskLRF without token subsampling and MaskLRF with subsampling. As described in Section III-C, token subsampling reduces computational cost by choosing only $t$ tokens as attention targets for each query token. Without token subsampling, MaskLRF causes an out-of-GPU-memory error at the finetuning stage. This is because the relative-pose aware self-attention using all the $N_p$ tokens does not fit to the GPU memory. On the other hand, MaskLRF with token subsampling shows computational efficiency competitive to the other algorithms both for pretraining and finetuning. However, MaskLRF with token subsampling requires more than twice the computation time compared to Point-MAE. Further improvement in the efficiency of MaskLRF should be considered to apply it to pretraining on datasets larger than ShapeNetCore55.

Table XIII. Comparison of computational efficiency for pretraining on ShapeNetCore55 and finetuning on SONN OBJ_BG.

| Methods | Pretraining | | Finetuning | |
|---|---|---|---|---|
| | Time per epoch [s] | GPU memory [GBytes] | Time per epoch [s] | GPU memory [GBytes] |
| Point-MAE | 55.6 | 20.0 | 8.13 | 11.5 |
| Point-M2AE | 286.2 | 21.4 | 20.14 | 19.6 |
| PointGPT | 75.8 | 11.1 | 9.86 | 11.1 |
| RIConv++ & SDMM | 924.8 | 14.3 | 15.61 | 3.9 |
| PaRot & SDMM | 1121.2 | 23.9 | 15.59 | 19.3 |
| MaskLRF without token subsampling | 282.3 | 14.4 | – | – |
| MaskLRF with token subsampling | 139.5 | 11.3 | 16.93 | 11.6 |

## V. CONCLUSION AND FUTURE WORK

### A. CONCLUSION

This paper tackled, for the first time, a rotation-invariant (RI) framework of self-supervised pretraining for 3D point set analysis. The proposed Masked Point Modeling (MPM) algorithm, called MaskLRF, learns RI latent shape features via masked autoencoding of local patches whose rotations are normalized by their Local Reference Frames. MaskLRF refines rotation-normalized patches by using self-attention layers with relative pose encoding, which can consider mutual pose differences among the patches. The pretext task that requires to reconstruct 3D grid-structured descriptors having rich 3D geometry facilitates to learn expressive latent features. The efficacy of MaskLRF was verified via extensive experiments on various downstream tasks. In addition, the in-depth evaluation validated the design of the proposed algorithm. These experiments revealed that MaskLRF is capable of learning rotation-invariant and highly generalizable latent 3D shape features in a self-supervised learning framework. More specifically:

- Although MaskLRF uses synthetic 3D point sets derived from 3D CAD models for pretraining, the learned latent features are effective in analyzing real-world (noisy) 3D point sets with inconsistent orientations, as shown in the experiments on real-world object classification and scene registration.

- MaskLRF performs well in scenarios where a small number of labeled 3D point sets are available for finetuning, as shown in the experiment on few-shot classification.

- The 3D shape features learned by MaskLRF are useful not only for object-level analysis (i.e., shape classification), but also for point-level dense prediction tasks such as part segmentation and scene registration.

- MaskLRF can learn robust latent features even if there are domain gaps (i.e., synthetic/realistic gap and

orientational gap) between training and evaluation data, as shown in the experiment on domain adaptation.

*B. FUTURE WORK*

This paper proposes the MaskLRF algorithm as the first attempt of the RI MPM framework. However, there is much room for further exploration within the RI MPM framework. Possible future directions include, for example,

- **Using better rotation invariance acquisition methods.** MaskLRF uses the traditional PCA-based LRF to normalize the orientation of local regions. However, the PCA-based LRF is sensitive to noisy 3D points. Therefore, the current MaskLRF may not fulfil its potential in analyzing real-world 3D point sets. To obtain rotation invariance in a more robust way, the use of learning-based LRF (e.g., [43], [47]) or feature extraction DNNs having rotation equivariance (e.g., [38], [39]) should be considered.

- **Pretraining with large-scale and realistic datasets.** The current MaskLRF uses nearly 50,000 synthetic 3D point set data for pretraining. Although its effectiveness was confirmed in the experiments, pretraining with a larger number of data samples may lead to better latent shape features. In addition, pretraining on realistic (not synthetic) 3D point set data is expected to gain robustness against 3D shapes having noisy 3D points, missing parts, and non-uniform point density. However, such a larger-scale pretraining would require improving the computational efficiency of MaskLRF.

- **Evaluation using more diverse downstream tasks.** The tasks of 3D point set analysis are not limited to classification, part segmentation and scene registration dealt with in this paper. We would like to verify the practicality of the proposed algorithm by evaluating it on additional downstream tasks such as object detection, scene segmentation and shape reconstruction.